# Generative AI in Map-Making: A Technical Exploration and Its Implications for Cartographers

Claudio Affolter
Institute of Cartography and Geoinformation, ETH Zurich
Swiss Federal Laboratories for Materials Science and Technology, Empa
Switzerland

Sidi Wu*
Institute of Cartography and Geoinformation, ETH Zurich
Switzerland

Lorenz Hurni
Institute of Cartography and Geoinformation, ETH Zurich
Switzerland

Yizi Chen
Institute of Cartography and Geoinformation, ETH Zurich
Switzerland

## Abstract

Traditional map-making relies heavily on Geographic Information Systems (GIS), requiring domain expertise and being time-consuming, especially for repetitive tasks. Recent advances in generative AI (GenAI), particularly image diffusion models, offer new opportunities for automating and democratizing the map-making process. However, these models struggle with accurate map creation due to limited control over spatial composition and semantic layout. To address this, we integrate vector data to guide map generation in different styles, specified by the textual prompts. Our model is the first to generate accurate maps in controlled styles, and we have integrated it into a web application to improve its usability and accessibility. We conducted a user study with professional cartographers to assess the fidelity of generated maps, the usability of the web application, and the implications of ever-emerging GenAI in map-making. The findings have suggested the potential of our developed application and, more generally, the GenAI models in helping both non-expert users and professionals in creating maps more efficiently. We have also outlined further technical improvements and emphasized the new role of cartographers to advance the paradigm of AI-assisted map-making. The code and pre-trained models are available at https://github.com/claudaff/generative-ai-mapmaking/.

## CCS Concepts

• **Computing methodologies → Artificial intelligence**.

## Keywords

GenAI, map generation, cartography

## 1 Introduction

Maps are essential tools that help us understand and navigate the world. They play a critical role in various fields by visualizing spatial phenomena and patterns in a structured and informative way. However, map-making largely relies on specific Geographic Information System (GIS) and cartography software solutions and involves complex workflows that require domain expertise [54]. This not only limits accessibility for non-experts but can also be time-consuming for professionals, particularly when it comes to repetitive tasks, such as regularly updating maps to incorporate minor changes or adapting them to different languages. In recent years, advances in generative AI (GenAI), particularly in image generation using models like Stable Diffusion [40], have opened up new possibilities for automating and enhancing the map-making process. Leveraging the power of GenAI in map-making can potentially unlock various applications, such as 1) efficient, rapid, and accessible map creation, which allows for producing maps instantly with minimal manual effort and limited domain knowledge; 2) map standardization, which harmonizes diverse sources of geospatial data into a unified cartographic scheme; 3) map restoration and revitalization, which involves reconstructing damaged or missing information in historical maps or synthesizing modern data in a historical style to bootstrap information extraction from historical maps, thereby safeguarding their cultural heritage. This aspect is particularly challenging with traditional software, as recreating or preserving historical elements in a meaningful way requires intricate adjustments that GenAI can simplify and automate.

Current GenAI models demonstrate promising performance in rendering images based on textual instructions (i.e., text prompts) provided by the users. However, textual input alone is often imprecise and insufficient for controlling the spatial layout and semantic composition—both are crucial for maps. To address this, we propose a novel map generation method using *ControlNet* [59], which allows us to integrate explicit controls such as spatial layouts and semantics to guide the generation process. With input vector data, our model can produce maps in multiple styles, including both modern and historical, as specified by a text prompt. This is illustrated in Figure 1. Moreover, to enhance the accessibility and usability of the generation model, we have developed a web-based application for map generation. This application eliminates the need for specialized GIS or cartography software and reduces the workload of rendering maps, as users no longer need to manually assign a style scheme to input vector data each time. A demonstration of the application interface is shown in Figure 2.

To the best of our knowledge, this is the first work to explore both semantic and spatial control for generating high-quality maps using generative AI. To understand how such a system could reshape the map-making process, we conducted a user study with professional cartographers from national mapping agencies. Our investigation is guided by three key research questions: the fidelity and visual quality of the generated maps (**Q1**), the usability and potential of the interactive web-based application for professional workflows (**Q2**), and the broader implications of GenAI for the future of cartographic practice (**Q3**).

*Corresponding author: sidiwu@ethz.ch

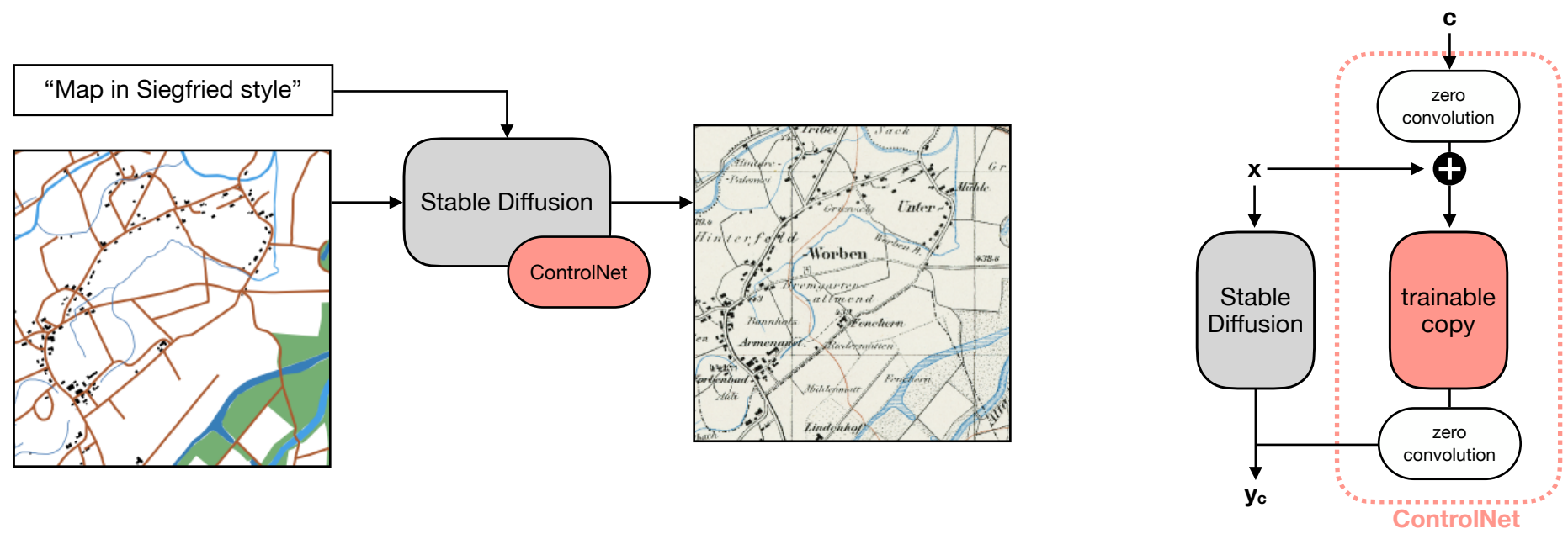

**Figure 1: Left: Overall concept. A text prompt specifying the map style, here *Siegfried* style, and vector data defining the spatial composition and semantic layout are provided as conditioning input to the *ControlNet*, built upon *Stable Diffusion*. Right: the basic blocks of *ControlNet*. $x$: feature maps from input image; $c$: feature maps from condition; $y_c$: output feature maps. It locks the original *Stable Diffusion*, creates a trainable copy, and connects them together using zero convolution layers, i.e., convolution layers with values initialized as zeros.**

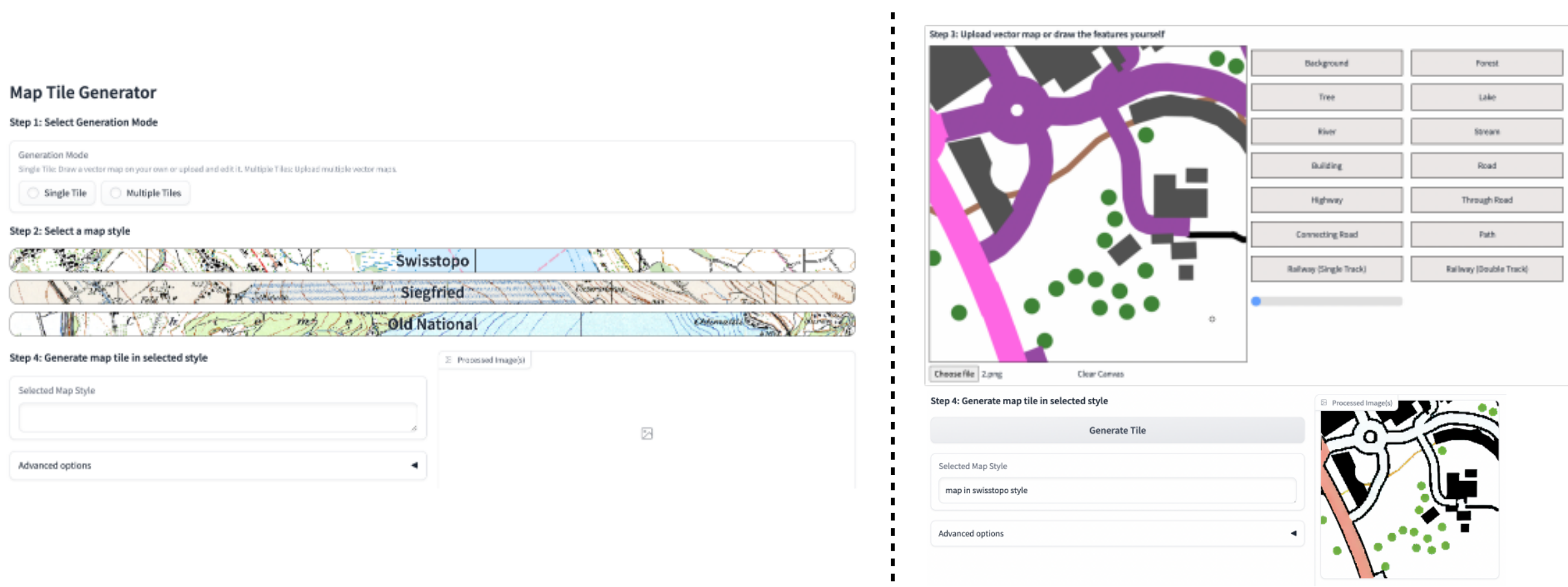

**Figure 2: Left: Web application interface. Detailed steps are explained in Section 3.5. Right: A detailed view of *Step 3*, where the control image is displayed, and *Step 4*, where a generated map tile in *Swisstopo* style is shown.**

## 2 Related work

Image generation is a core research area in the field of computer vision. Generative models have evolved significantly over the past decade, beginning with Generative Adversarial Networks (GANs) [20, 62], Variational Autoencoders (VAEs) [27, 28], and autoregressive models [47], to the more recent diffusion models [17, 23]. These generative models have been applied in a wide range of applications, including generating realistic images and videos [23, 39, 48], artistic or stylized portraits [31, 57], translating images into different visual styles [24, 43, 62], super-resolution [29, 45], inpainting [32, 58], text-to-image synthesis [38, 40], and conditional generation based on semantic maps or sketches [11, 36].

More recently, diffusion models have demonstrated superior performance in producing photorealistic images, gaining wide adoption in both research and industry. They not only achieve superior quality compared to GANs [18], but also suffer less from the mode collapse, notorious for training GAN models [46]. Image diffusion models have been widely used to generate images with textual prompts [35, 38, 40, 42, 44]. Since in many cases, text prompts are not sufficient to control image diffusion models, additional prompts in other forms have been explored. For example, some works add segmentation maps to highlight regions of interest or control the semantic layout [2, 4, 16, 50]. Other techniques include sketch guidance [7, 15, 49], bounding boxes [10], and depth maps [8], to control the geometry in the generation. Moreover, layout-to-image generation has been introduced to enable the creation of complex scenes containing multiple objects [14, 56, 61]. In addition, methods supporting multiple external control signals have been developed [5, 21, 30, 34, 60]. Among these, *ControlNet* [59] has become one of the most widely used methods for controlling image diffusion models. *MultiDiffusion* [6], unifies several controls over the generated content by binding together multiple diffusion generation processes. Closely related to this work, *SatDM* [3] controls diffusion models using building footprints for generating synthetic satellite imagery. Additionally, the potential of rich-text editors as an alternative to plain text prompts for enabling more precise control of text-to-image synthesis has been investigated [19].

For map generation, existing works have explored the translation from satellite images to maps [12, 25, 51]. [26] made the first attempt using the pre-trained DALL·E 2 [38] to generate maps by specifying map type, region, country, and other descriptions, with textual prompts.

## 3 Methodology

### 3.1 Preliminary: Stable Diffusion and ControlNet

Diffusion models learn to synthesize data by reversing a gradual noising process. During training, Gaussian noise is incrementally added to images over many steps, and a neural network is trained to denoise them step by step. At inference, new images are generated by iteratively denoising a random noise sample. *Stable Diffusion* is a latent diffusion model (LDM) that performs image generation by denoising in a compressed latent space rather than pixel space, improving efficiency while maintaining quality [40]. It combines a variational autoencoder (VAE) for latent encoding, a U-Net denoising network, and a text-conditioning mechanism (e.g., CLIP [37]). During training, Gaussian noise is added to latent representations, and the model learns to reverse this process, conditioned on corresponding text embeddings. At inference, images are generated by iteratively denoising a noise sample in latent space, guided by a textual prompt as the condition.

Built upon *Stable Diffusion*, *ControlNet* [59] injects additional conditions to guide the diffusion process, as shown in the right part of Figure 1. To achieve this, the parameters of the original *Stable Diffusion* blocks are frozen (i.e., locked to prevent them from being updated during training). The locked parameters preserve the capabilities of the pre-trained image diffusion model. Simultaneously, these blocks are cloned to create a trainable copy (with unfrozen weights) that takes external conditioning as input (e.g., sketches, normal maps, segmentation maps). The feature maps from the input image $x$ are passed through the original frozen blocks, while the features from the conditional input $c$, together with the input features, are processed by a trainable branch. The two branches are connected through zero-convolution layers to generate the output features $y_c$, which are incorporated in the diffusion process. Zero-convolution layers are the 1×1 convolutions initialized with zero values, ensuring that the initial outputs are identical to the original *Stable Diffusion* model. These layers allow the model to gradually learn how to integrate the conditioning information without disrupting the original generation process. Additionally, a shallow network of four convolution layers is trained to encode the condition from image space to feature space. The text prompt is encoded with the pre-trained CLIP text encoder [37], consistent with the standard *Stable Diffusion* model, and is passed to the frozen blocks to guide the generation style.

### 3.2 Data

In our work, we use three different map styles covering both modern and historical periods: 1) *Swisstopo* style—the style used in the contemporary national Maps of Switzerland from the Federal Office of Topography (*Swisstopo*), the Swiss national mapping agency[1]; 2) *Siegfried* style—the style used in the topographic map series *Siegfried Maps* published between 1879 and 1949; 3) *Old National* style—the style used in the nation-wide map series *Old National Maps*, published between the 1950s and 1990s. The maps from both *Siegfried Maps* and *Old National Maps* were scanned and archived by *Swisstopo*. All the maps are at the scale of 1:25 000 with a spatial resolution of 1.25m/pixel. Each map sheet in *Swisstopo* style and *Old National* style has the size of 14 000 × 9 600 pixels while each map sheet in *Siegfried* style has the size of 7 000 × 4 800 pixels. We use an average of 6 map sheets for training different map styles. We use the contemporary vector dataset *Vector 25* from *Swisstopo* at the scale of 1:25 000 corresponding to the same location as our additional condition to the *ControlNet* model. It contains the vectors of different geographical features such as hydrology, roads, buildings, and vegetation.

### 3.3 Data processing

We prepared the dataset for each map style following a triple structure specified by *ControlNet*: the *source control* vector data in the form of raster images with class labels assigned to each pixel, the *target* raster map, as well as the *text prompt* that describes the map style (e.g., "map in *Swisstopo* style"). To rasterize the source vector data, we assigned class labels to each pixel depending on the object categories (e.g., roads, rivers, buildings) and adjusted line widths and point sizes to match the corresponding features in the raster maps. We adapted the object categories slightly for historical maps (i.e., *Old National* and *Siegfried* styles), as their categories do not fully align with the modern vector data. Table 2 provides an overview of the class labels present in each of the three map styles. Furthermore, because diffusion models often produce illegible or unrealistic text, we generated binary masks for all textual labels. These masks were applied to both the raster maps and the rasterized vector data, so that the text regions were blanked out entirely. This prevented the model from learning to reproduce text. At inference, as no label masks were applied to the rasterized vector data, the model would not hallucinate any features corresponding to text labels. We call this strategy the *mask supervision.* For *Swisstopo* style, creating text masks was straightforward as we had annotation vectors (bounding boxes) that corresponded directly to the locations of text labels. For the historical styles, we used Keras-OCR [33], a pre-trained optical character recognition (OCR) model, to detect text labels.

We divided each map sheet into smaller tiles of 512×512 pixels for training. We observed that using the original scale of 1:25 000 often led to implausible generation of small objects and blurry outputs, likely due to excessive information density. To address this, we upsampled the map sheets by a factor of 5 before tiling, allowing the model to better capture fine-grained spatial details. In total, we created three training datasets consisting of 145 726 tiles for *Swisstopo* style, 62 928 tiles for *Siegfried* style, and 87 841 tiles for *Old National* style. For evaluation, we created three test sets, consisting of 100 tiles for each map style. The use of a relatively small test dataset is due to a trade-off with inference time, which is discussed in detail in the following section.

[1] https://www.swisstopo.admin.ch

### 3.4 Model training and inference

We merged the three datasets, resulting in 296 495 training tiles in total, to train a *ControlNet* model capable of generating map tiles in all three styles. Training was performed on three Nvidia Tesla A100 GPUs (3 × 40 GiB) for a total duration of 36 hours (around 20 000 steps). The batch size was set to 10 per GPU, with a constant learning rate of 2e-6 and AdamW optimizer. Moreover, we trained *ControlNet* with the additional "sd_locked" parameter set to *False*, unlocking certain layers in *Stable Diffusion*.

At inference time, we observed that random seed initialization did not significantly influence the generation results for the *Swisstopo* and *Old National* styles. By contrast, the outputs for the *Siegfried* style showed substantial variation depending on the selected seed. This may be explained by a significant discrepancy between the source vector and the target maps, due to the large temporal gap, over a century between them. To avoid manual inspection and seed selection, we proposed an automated strategy. We first trained a semantic segmentation model using U-Net [41], which is commonly used for historical map segmentation [13, 22, 52, 53, 55], to segment the generated maps. We then compared the mean Intersection over Union (mIoU) between the segmentation results from the generated maps and the corresponding rasterized source vector images. Additionally, we observed that the generation model tended to produce noisy (fake) labels in the background areas of *Siegfried*-style maps. This was likely due to Keras-OCR failing to mask all text labels during pre-processing. To quantify this effect, we computed the standard deviation of pixel values in the background regions (Std_background) as a measure of background noise. Therefore, in total, we combined both mIoU and Std_background as our criteria to automatically select the best seed initialization, after generating multiple outputs using several different random seeds. On average, generating 100 test images took approximately 25 minutes for the *Swisstopo* and *Old National* styles. For the *Siegfried* style, however, inference time increased by a factor of four due to the seed selection strategy, where six random seeds were evaluated per sample.

### 3.5 Application development

We have developed a web application (see Figure 2) for generating map samples using Gradio [1], with our trained *ControlNet* serving as the backend model. The map generation process is carried out in four steps:

- *Step 1*: The user selects the generation mode, either "Single Tile" or "Multiple Tiles". For "Multiple Tiles", all tiles will be generated in a batch and packaged for download.
- *Step 2*: The user chooses a desired map style by clicking one of three predefined style buttons.
- *Step 3*: The user uploads vector map tiles, pre-processed as raster semantic maps (in .png format), which are displayed on a canvas. These control images can be edited directly, or users can create them from scratch using predefined strokes, with adjustable pen size for each feature.
- *Step 4*: The user clicks the "Generate Tile" button to start the map generation process. The resulting map is displayed on the right side of the interface.

## 4 User study

### 4.1 Participants and materials

We conducted a user study with eleven participants (three females and eight males) from national cartography agencies, the *Swiss World Atlas*[2] and *Atlas of Switzerland*[3]. Their ages ranged from 21 to 66 years, with a median age of 28. They had a median of 7 years of academic and professional experience in cartography. Eight participants reported having little to no prior experience with text-to-image generation tools. We prepared a printed questionnaire to be filled out by each participant. The questionnaire is comprised of four parts: the demographic information and experience with GenAI, map fidelity and quality assessment (**Q1**), usability test of the web-application (**Q2**), and an interview about the further implications of GenAI in map-making (**Q3**).

### 4.2 Questions and tasks

**Map fidelity assessment (Q1)**. We created and printed three assessment sheets—one for each map style. Each sheet contained three tasks, mixing real and synthetic map tiles in various configurations:

- *Task 1 (direct comparison)*: Two pairs of medium-sized maps are displayed, each pair consisting of the real map and the synthetic replica based on the corresponding vector input. The user has to determine for each pair which map is the real one.
- *Task 2 (single assessment)*: Ten small map tiles from different regions, either real or synthesized, are mixed. The user has to determine for each tile whether it is real or not.
- *Task 3 (mixed assessment)*: A medium-sized map is shown, stitching both real and synthetic tiles into a complete map snippet. The user has to identify for each tile if it is real or not.

An example of **Q1** for *Swisstopo* style can be found in Figure 5. After finishing the tasks, the users were required to rate the overall similarity between the real and generated maps for each map style, on a scale from 0 (completely dissimilar) to 5 (identical), with an explanation of what helped them to distinguish. While all participants completed the tasks in the same sequence (*Task 1, 2 and 3*) the style order (i.e., the order in which they received the assessment sheets) varied. No time limit was set.

**Usability test (Q2)**. For the usability test, we ran our developed web application on a 13" Apple MacBook Air using the Google Chrome web browser. We have asked the participants to finish three tasks with our web application, without any prior introduction or tutoring:

- *Task 1*: Given a control raster semantic map, the user should generate a map tile in *Swisstopo* Style.
- *Task 2*: The user has to draw an control image on the canvas using at least three feature classes/colors, and use it to generate a map tile in *Siegfried* Style.

[2] https://www.schweizerweltatlas.ch/
[3] https://www.atlasderschweiz.ch/

- *Task 3*: The user is required upload multiple control raster semantic maps at once, conduct the batch map generation in *Old National* Style, and then download the resulting map images.

Afterward, the participants were asked to complete a System Usability Scale (SUS) [9], a standard questionnaire to test a system's usability. They also provided general feedback and suggestions for improvement, as well as potential use cases for the developed tool.
**Interview (Q3)**. In the final part of the study, a post-study interview was conducted. Participants were asked a series of open-ended questions about the potential applications of AI-generated maps, both in general and in relation to their daily work, as well as suggestions for improvements to develop a cutting-edge map generation tool. They also shared their views on ethical and practical concerns, and reflected on the future role of cartographers in light of the rapid advancement of generative AI. Their responses were documented in real time.

## 5 Results

### 5.1 Map generation

Figure 3 illustrates the performance of our trained ControlNet model on each test set. The generated map tiles were stitched together to create 5 120 × 5 120 pixel output maps. In the left column, the control images (i.e., stitched input vector map tiles) are displayed, while the second and third columns present the corresponding output maps generated by our model, both before and after post-processing.

For the *Swisstopo* style, post-processing involved minor automatic color corrections for background, river, and building features to improve visual consistency between adjacent tiles. Since this map style originates from purely digital sources, we were able to extract a consistent color palette for these features and apply it to the generated tiles for automatic color correction. In contrast, the *Old National* maps are scanned reproductions of original analogue maps, which often exhibit inherent color variations for features from different maps. As a result, a single, consistent color palette is not feasible to obtain. Therefore, post-processing for this style was limited to homogenizing only the background color to improve visual coherence.

For *Siegfried* style, we augmented the predictions with different seed options and selected the best results, as discussed earlier in Section 3.4. Furthermore, as we observe a significant shift of contour lines between *Siegfried* maps and the source vector due to the low spatial accuracy of the historical height profile, we omitted this class during training. Instead, we projected the contour lines with brown strokes in the final output during post-processing.

Figure 3 shows qualitative results of our model across the three styles, *Swisstopo*, *Old National*, and *Siegfried*. The right-most column presents the real maps centered at the same location, originating from contemporary times, the 1980s, and the 1940s, respectively. The results show that, while the raw generations already exhibit strong visual similarity to the reference maps, subsequent post-processing significantly enhances visual quality and consistency—particularly for hydrological features—thereby improving the overall usability of the generated maps.

To provide more insight into the map generation process, Figure 4 shows selected image log entries from *ControlNet*, which we used to monitor the model's training process. The visual logs reveal a clear transition—from generating maps in arbitrary, uncontrolled styles to producing outputs that increasingly adhere to both the semantic layout and the specified cartographic style. Notably, we observed the so-called *sudden convergence phenomenon* [59], which occurred between 2 700 and 3 000 training steps, marking a sharp improvement in generation quality and control. We continued to train until around 20 000 steps to further enhance generation fidelity and to suppress unwanted artifacts, such as residual text labels, guided by our *mask supervision* as described in section 3.3.

### 5.2 Findings of the user study

**Map fidelity assessment (Q1)**. Table 1 summarizes the quantitative results related to **Q1**. A map tile or snippet is considered a true positive if it is correctly identified as real, and a true negative if it is correctly identified as synthetic. Based on these classifications, we compute the average precision, recall, and F1 score. Additionally, we report the average similarity score assigned by participants, reflecting the perceived resemblance between real and synthetic maps on a scale from 0 (completely dissimilar) to 5 (identical).

Among the three tasks, participants found the direct comparison in *Task 1* easiest, consistently distinguishing real from synthetic maps across all styles. In contrast, the single-tile evaluation in *Task 2* and the mixed assessment in *Task 3* proved more challenging, especially for the *Old National* style, and even more so for the *Swisstopo* style, where performance approached random guessing. This suggests that while generative artifacts are detectable when synthetic maps are directly compared to their real counterparts, they become less apparent when tiles are assessed in isolation, likely due to the visual complexity and irregularity of maps. Furthermore, visual cues like consistencies, spatial continuities and other contextual information between adjacent tiles can also explain the ease of detection in larger map snippets compared to individual tiles. The struggle for *Swisstopo* style suggests that this style exhibits particularly high visual quality and fidelity in its generated outputs. This can also be suggested by the highest similarity score of *Swisstopo* style. Interestingly, although the *Siegfried* style shows generally lower visual quality and semantic accuracy (with the average IoU of only 0.3 compared to the ground truth semantics), participants were often unable to correctly identify synthetic tiles. Many attributed this to the inherent imperfections in historical maps, which blur the boundary between real and generated content.

Further qualitative analysis of participants' reasoning revealed common indicators for detecting synthetic maps. For the *Swisstopo* style, these included subtle color inconsistencies, cartographic errors (e.g., distorted symbols), broken lines, minor tile misalignments at borders, and topological inaccuracies. In the *Old National* style, participants noted uneven line widths, clear tile border misalignments, implausible color variations, discontinuous features, incorrect filage (coast-parallel lines), and inconsistent geometries. For the *Siegfried* style, detection was typically based on severe tile misalignments, erroneous textures in rivers and forests, discontinuous roads, incorrect symbology for roads and railways, missing details (such as hachures or embankments), and overly vibrant colors.

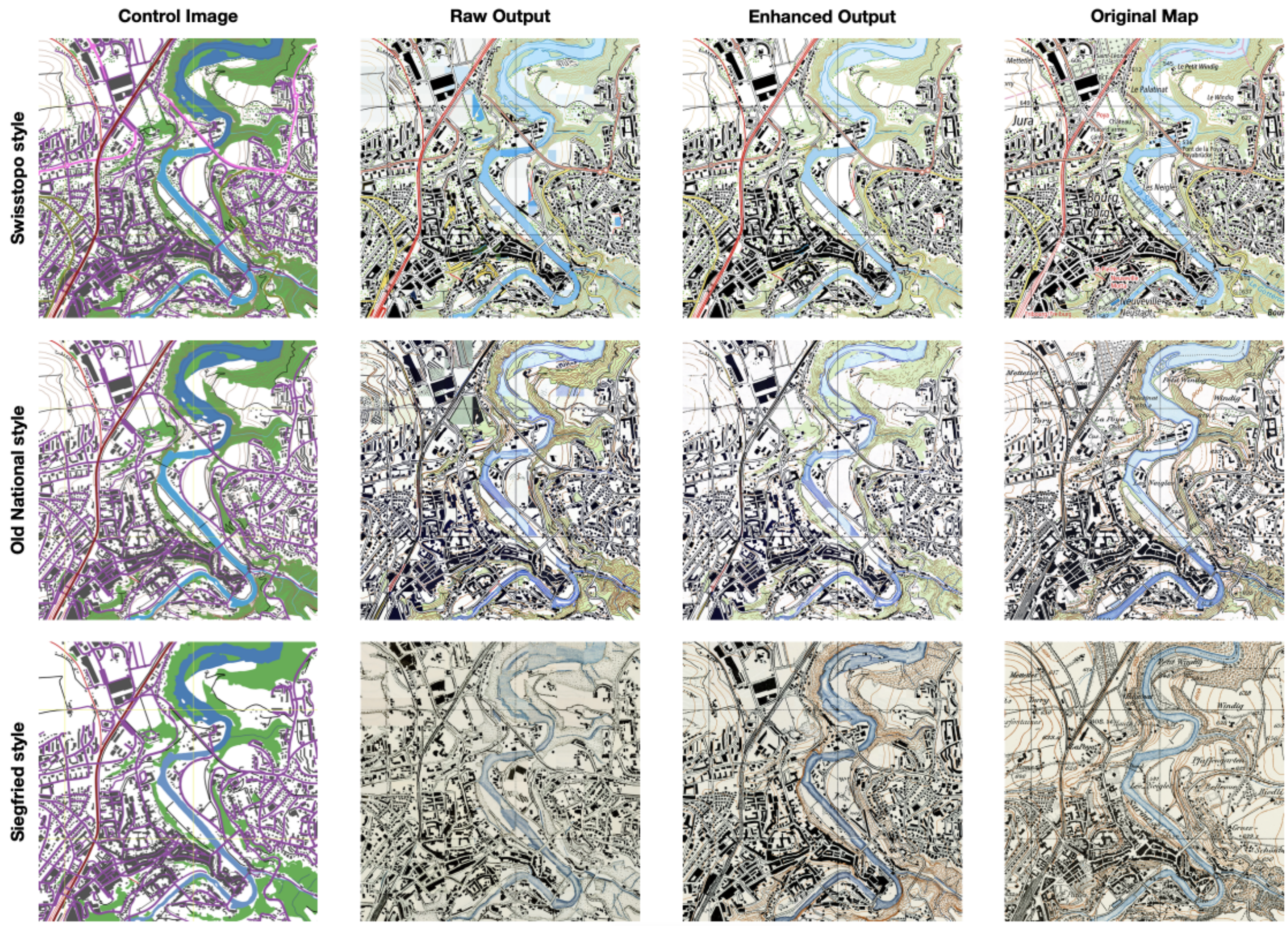

**Figure 3: Qualitative performance of our trained ControlNet model on the test sets. From left to right: control images, raw output, enhanced output after post-processing, the real map depicting the same location from contemporary times for *Swisstopo* style, the 1980s for the *Old National* Style, and the 1940s for the *Siegfried* style, respectively.**

**Table 1: Quantitative results of map fidelity assessment. For each map style, we report the average accuracy, precision, and F1 score across all three tasks, along with the overall similarity score rated by participants.**

| | *Swisstopo* | | | *Old National* | | | *Siegfried* | | |
|---|---|---|---|---|---|---|---|---|---|
| Metric | Task 1 | Task 2 | Task 3 | Task 1 | Task 2 | Task 3 | Task 1 | Task 2 | Task 3 |
| Precision | 1.00 | 0.58 | 0.54 | 1.00 | 0.98 | 0.91 | 0.95 | 0.69 | 0.59 |
| Recall | 0.95 | 0.51 | 0.68 | 1.00 | 0.73 | 0.75 | 0.95 | 0.75 | 0.84 |
| F1 | 0.97 | 0.54 | 0.60 | 1.00 | 0.84 | 0.82 | 0.95 | 0.72 | 0.69 |
| Similarity | 4.14 | | | 3.11 | | | 3.61 | | |

**Usability test (Q2)**. In this study, we have obtained an average SUS score of 82.73 ($\sigma$ = 10.27), given the three tasks provided, indicating a high level of usability and user satisfaction. Participants particularly appreciated the application's simplicity and the speed of generation. In terms of general feedback, several participants suggested further improvements such as more intuitive and standardized icons, clearer color schemes, helpful tooltips, and expanded functionality, especially to visualize the map as a whole for the "Multiple Tiles" generation mode. More than half of the participants indicated their preference for the system to be integrated into a GIS platform, e.g., as a plugin, to better align with standard cartographic workflows, while the rest favored its current independence from any specific GIS software. Additionally, several participants suggested enabling direct input of original vector formats (e.g., shapefiles) and handling the rasterization process automatically on the backend, to reduce the pre-processing step. Overall, participants recognized strong potential for this application across a wide range of use cases, particularly for users seeking to generate a topographic map quickly without extensive cartographic expertise, including teachers, students, historians, graphic designers, urban planners, game developers, and geographers.

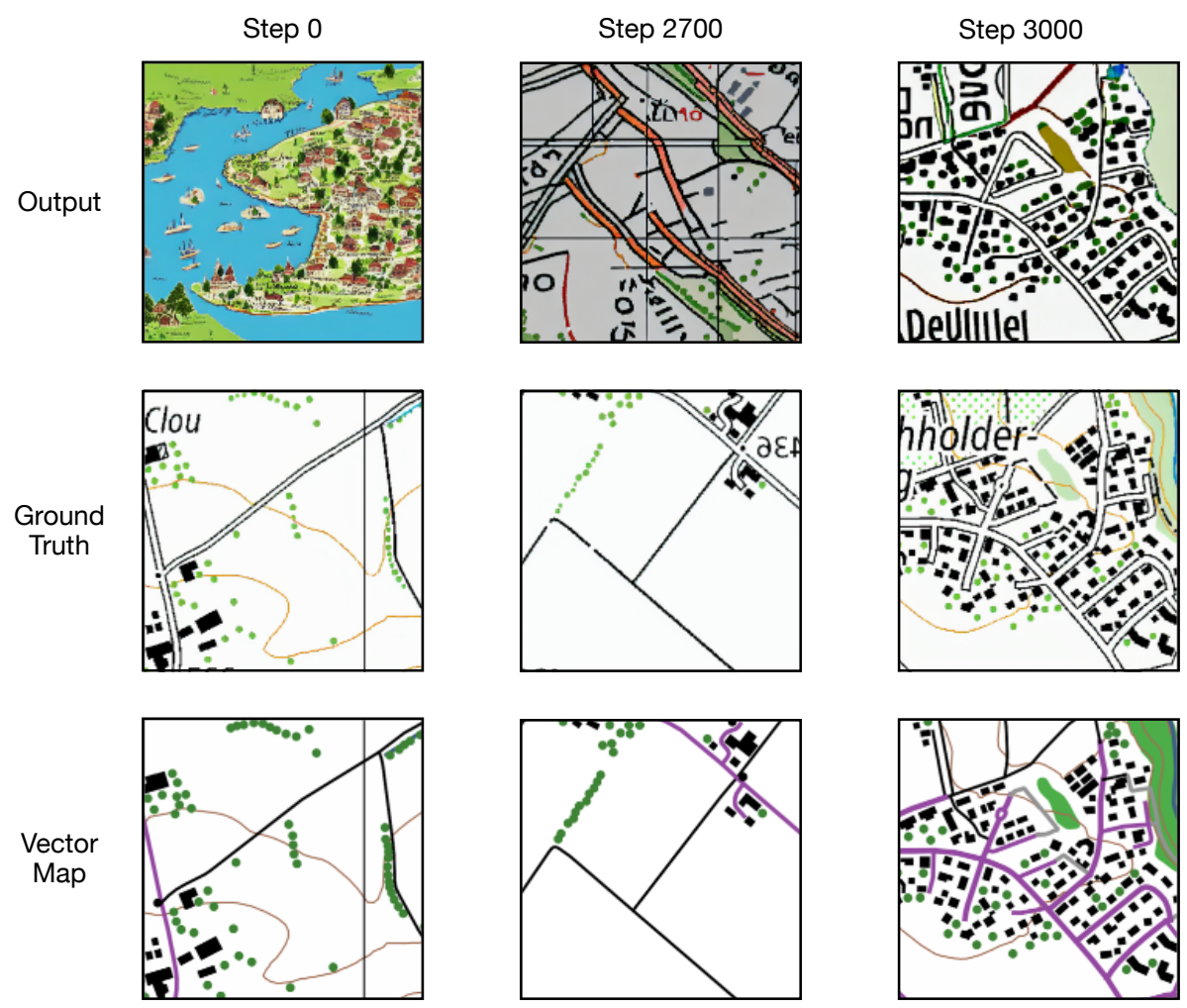

**Figure 4: ControlNet image log. At predefined intervals (e.g., every 300 steps), the model outputs generated images (top row) alongside the corresponding ground truth (target) and vector map tiles (source control).**

**Interview (Q3)**. Here we summarize the findings of our post-study interview about participants' reflections on AI-generated maps. Participants identified a wide range of potential applications, both for laypersons and within their professional cartographic workflows. These included updating maps in a consistent visual style, rendering vector data in multiple styles more efficiently, simulating spatial scenarios—such as proposed developments or land use changes—in different cartographic styles, rapidly generating maps for discussions, tailoring base maps to specific tasks, and synthesizing training data for machine learning models. To transform the system into a state-of-the-art GenAI tool for cartographers, the participants emphasized the need to improve the underlying model to achieve higher output quality, both in terms of colors and textures, as well as adherence to cartographic principles (such as symbolic consistency, generalization, and balancing spatial composition). Participants emphasized that accuracy is a critical requirement when generating maps in modern cartographic styles intended for professional use, whereas artistic expression and visual appeal are often more important for historical map styles.

A majority (73%) raised ethical and practical concerns about fully trusting or relying on AI-generated maps. Key concerns included incorrect topology, implausible spatial accuracy, and missing or distorted features, especially problematic in maps intended for orientation or navigation. Participants also warned that misplaced or inaccurately rendered political borders could cause confusion or even political disputes. While they appreciated the efficiency and creative potential GenAI offers, participants also acknowledged the possibility that its adoption may reduce the demand for some traditional cartographic roles. Nonetheless, they unanimously agreed that GenAI should be seen not as a replacement but as a collaborative tool. Cartographers will continue to play a vital role in verifying and refining the output, ensuring that cartographic conventions and quality standards are properly upheld.

# 6 Discussion

## 6.1 The map-generation model

As we can see from the last section, the model works best for the *Swisstopo* style, where even experienced cartographers struggled to consistently identify synthetic map tiles, thus, our model can be potentially used to update the modern maps when changes happen, to simulate the new city layout for spatial planning in response to a planned change/development, or to show non-contemporary data in the modern style for better comparison with modern data. By comparison, the generations in the *Old National* and *Siegfried* styles are less immediately usable and require post-processing to achieve acceptable quality—particularly for the *Siegfried* style, which involves substantial processing time. This limitation is likely due to misalignments between the modern vector data and the older map references used during training. More carefully curated and temporally aligned datasets could potentially improve output quality. Given the current time demands, applying the model at a large scale for historical styles is impractical and may be best suited for use cases that require only rough historical sketches for illustrative purposes, where spatial accuracy and high generation fidelity are not essential.

In **Q1**, participants highlighted issues such as incorrect or inconsistent geometry and topology in the generated maps, as well as noticeable color and style mismatches between adjacent tiles in large-scale outputs. To address these limitations, future map generation models should aim to improve both geometric and topological accuracy, as well as respect the cartographic principles. Since current models generate tiles independently, stylistic inconsistencies can arise when tiles are stitched together, decreasing overall visual coherence. Therefore, models capable of generating coherent sequences of tiles should be developed. Additionally, expanding the range of available map styles and incorporating more descriptive textual prompts for finer style control would further improve the flexibility and usability of the model.

## 6.2 The application system

In **Q2**, cartographers appreciated the simplicity and responsiveness of our system, as reflected in the high SUS score. However, they noted that the current prototype is primarily suitable for generating quick sketches or previews—such as illustrating concepts in discussions or visualizing vector data in different styles—and is not yet adequate for professional applications requiring high visual quality and spatial accuracy. Addressing this gap requires not only improvements to the underlying generative models but also better integration with existing cartographic software and workflows, along with expanded functionality. While the integration of the system into GIS platforms improves its alignment into professional workflows, it also limits the accessibility for non-expert users without specialized software. To balance these needs, future development could consider offering separate versions tailored to professional and general users.

### 6.3 Implications of GenAI to Cartographers

In **Q3**, the interviewed cartographers expressed an optimistic outlook about the role of GenAI in map-making. They acknowledged its potential, especially for laypersons without domain expertise, for educational, illustrative, or creative purposes. At the same time, they also recognized it as a convenient tool for professionals to illustrate concepts and prototype design ideas quickly, to reduce repetitive work such as small map updates, and to generate customized maps for specific tasks with reduced manual effort. However, they expressed their ethical and practical concerns about fully trusting the maps created by GenAI. Thus, they have stressed the necessity of cartographers in the validation, refinement, and approval of AI-generated content. Ultimately, they viewed GenAI not as a replacement for cartographers but as a collaborative partner capable of handling repetitive or low-stakes tasks and even inspiring creativity—an approach we can refer to as "AI-assisted map-making". The role of cartographers may shift from a manual producer to a critical reviewer, editor, supervisor, and curator of AI-generated maps, to ensure the outputs meet standards of cartographic quality, usability, and ethical responsibility.

## 7 Conclusions

In this paper, we presented a technical exploration of applying one of the GenAI models *ControlNet* to generate maps in three distinct styles, both modern and historical, using the vector data to control the spatial layout and semantic composition, and simple textual prompts to define the generated style. We have developed a web-based application integrating the model at the backend to make this tool independent from GIS software and widely accessible. We conducted a user study with professional cartographers from national mapping agencies to evaluate the fidelity of generated maps, the usability of the web application, and further implications of GenAI for cartographers. Through both the user study and visual inspection, we find that our current model shows great potential in generating high-quality and accurate maps in the modern style, while for historical maps, it's best suited for tasks for illustrative purposes that don't strictly demand accuracy and fidelity. Further improvements should focus on enhancing the geometric and topological accuracy of generated maps, ensuring stylistic consistency across adjacent map tiles, and curating higher-quality training datasets, especially for historical styles. Our web application has been recognized for its simplicity and quick response, yet it also requires additional functionality and integration with a professional cartographic workflow to support professional uses in the future. The interviewed cartographers expressed their optimism with caution about the role of GenAI for map-making. They acknowledged its value for both non-expert users and professionals, especially for accelerating repetitive tasks and inspiring creative workflows. The important implication of this cutting-edge technology for them is not a replacement, but a collaborative partner in the paradigm of AI-assisted map-making, where the role of cartographers may shift from manual production to critical review, refinement, and validation. Despite a likely reduction of traditional cartographic tasks, the domain knowledge is still essential to ensure the generated maps adhere to cartographic quality, usability, and ethical responsibility, thereby shaping and supervising the outputs of this emerging technology.

## A Swisstopo Style Assessment Sheet

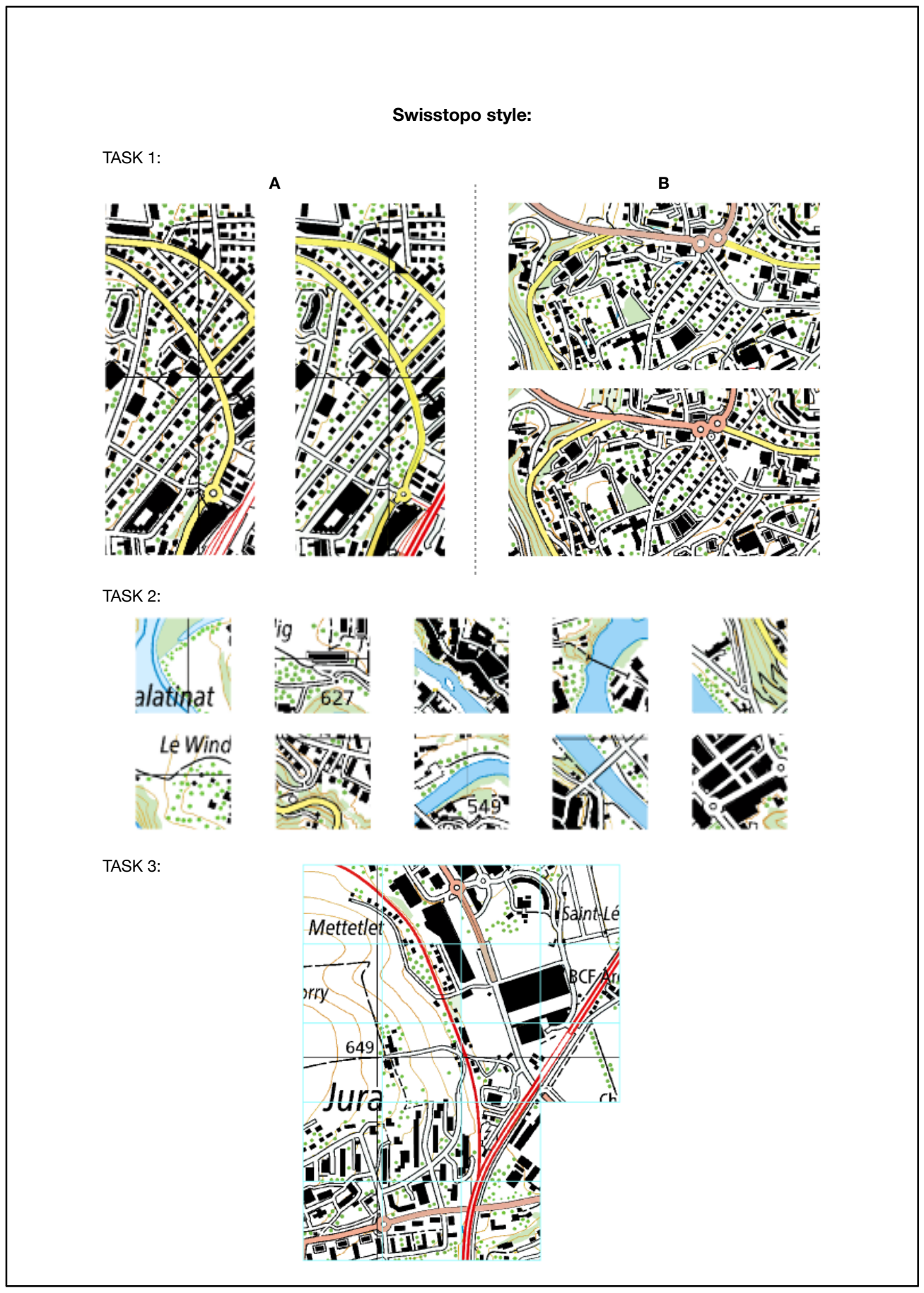

**Figure 5: Swisstopo Style Assessment Sheet: One of the three style assessment sheets used in the user study for the map fidelity assessment (Q1). The sheets for Old National and Siegfried style followed the same structure.**

## B Class labels

**Table 2: Presence of class labels in Swisstopo, Old National, and Siegfried style.**

| Class label | *Swisstopo* | *Old National* | *Siegfried* |
|---|---|---|---|
| Background | ✓ | ✓ | ✓ |
| Building | ✓ | ✓ | ✓ |
| Coordinate grid | ✓ | ✓ | ✓ |
| Railway (single track) | ✓ | ✓ | ✓ |
| Railway (multi track) | ✓ | ✓ | ✓ |
| Railway bridge | ✓ | | |
| Highway | ✓ | ✓ | |
| Highway gallery | ✓ | | |
| Road | ✓ | ✓ | ✓ |
| Through road | ✓ | | |
| Connecting road | ✓ | | |
| Path | ✓ | ✓ | ✓ |
| Depth contour | ✓ | ✓ | |
| River | ✓ | ✓ | ✓ |
| Lake | ✓ | ✓ | ✓ |
| Stream | ✓ | ✓ | ✓ |
| Tree | ✓ | | |
| Contour line | ✓ | ✓ | |
| Forest | ✓ | ✓ | ✓ |